\documentclass[10pt,twocolumn,letterpaper]{article}

\usepackage{cvpr}              %

\definecolor{cvprblue}{rgb}{0.21,0.49,0.74}
\usepackage[accsupp]{axessibility}  %
\usepackage[pagebackref,breaklinks,colorlinks,allcolors=cvprblue]{hyperref}
\usepackage{makecell}
\usepackage{multirow}

\def\confName{CVPR}
\def\confYear{2025}

\title{LSNet: See \underline{L}arge, Focus \underline{S}mall}

\author{Ao Wang$^{1}$ \quad Hui Chen$^{2}$\thanks{Corresponding author.} \quad Zijia Lin$^{1}$ \quad Jungong Han$^{3}$ \quad Guiguang Ding$^{1}$ \\
		$^1$School of Software, Tsinghua University \quad $^2$BNRist, Tsinghua University \\$^3$Department of Automation, Tsinghua University\\
		{\tt\small wanga24@mails.tsinghua.edu.cn \quad jichenhui2012@gmail.com \quad linzijia07@tsinghua.org.cn} \\ 
		{\tt\small jungonghan77@gmail.com \quad dinggg@tsinghua.edu.cn }}

\begin{document}
\maketitle

\begin{abstract}
    Vision network designs, including Convolutional Neural Networks and Vision Transformers, have significantly advanced the field of computer vision. Yet, their complex computations pose challenges for practical deployments, particularly in real-time applications. To tackle this issue, researchers have explored various lightweight and efficient network designs. However, existing lightweight models predominantly leverage self-attention mechanisms and convolutions for token mixing. This dependence brings limitations in effectiveness and efficiency in the perception and aggregation processes of lightweight networks, hindering the balance between performance and efficiency under limited computational budgets. In this paper, we draw inspiration from the dynamic heteroscale vision ability inherent in the efficient human vision system and propose a ``See Large, Focus Small'' strategy for lightweight vision network design. We introduce LS (\textbf{L}arge-\textbf{S}mall) convolution, which combines large-kernel perception and small-kernel aggregation. It can efficiently capture a wide range of perceptual information and achieve precise feature aggregation for dynamic and complex visual representations, thus enabling proficient processing of visual information. Based on LS convolution, we present LSNet, a new family of lightweight models. Extensive experiments demonstrate that LSNet achieves superior performance and efficiency over existing lightweight networks in various vision tasks. Codes and models are available at \url{https://github.com/jameslahm/lsnet}.
\end{abstract}

\section{Introduction}
\label{sec:intro}
Vision network designs have consistently been a focal point of research in the field of computer vision~\cite{han2022survey,he2016deep,dosovitskiy2020image,liu2022convnet,liu2021swin,zhang2024vision}, where two prominent network architectures, \ie, Convolutional Neural Networks (CNNs)~\cite{lecun1998gradient,krizhevsky2012imagenet,he2016deep,liu2022convnet,hou2024conv2former} and Vision Transformers (ViTs)~\cite{dosovitskiy2020image,touvron2021training,liu2021swin,yu2022metaformer,wu2022p2t,qian2022makes}, have significantly pushed the boundaries in various computer vision tasks~\cite{he2015spatial,wang2021pyramid,sun2023vicinity,yao2023dual,wang2024yolov10,chen2020imram,chen2018show}. However, both of them have traditionally been computationally expensive, presenting remarkable challenges for their practical deployments, especially for real-time applications~\cite{liu2023efficientvit,li2022efficientformer}.

Recently, researchers have been actively exploring the lightweight and efficient designs of vision networks~\cite{chen2022mobile,pan2022edgevits,mehta2021mobilevit,maaz2022edgenext,huang2023adaptive,vasu2023fastvit} for practical applications.
Despite effective, these lightweight models typically rely on certain basic modules, such as self-attention mechanism~\cite{vaswani2017attention,dosovitskiy2020image,wang2018non} and convolution~\cite{lecun1998gradient,krizhevsky2012imagenet}, for token mixing~\cite{tolstikhin2021mlp}. This reliance poses challenges regarding the efficiency and effectiveness of the underlying \textit{perception} and \textit{aggregation} processes within lightweight networks, often compromising the architectural expressiveness or inference speed.

\begin{figure*}[t]
  \centering
  \includegraphics[width=0.84\textwidth]{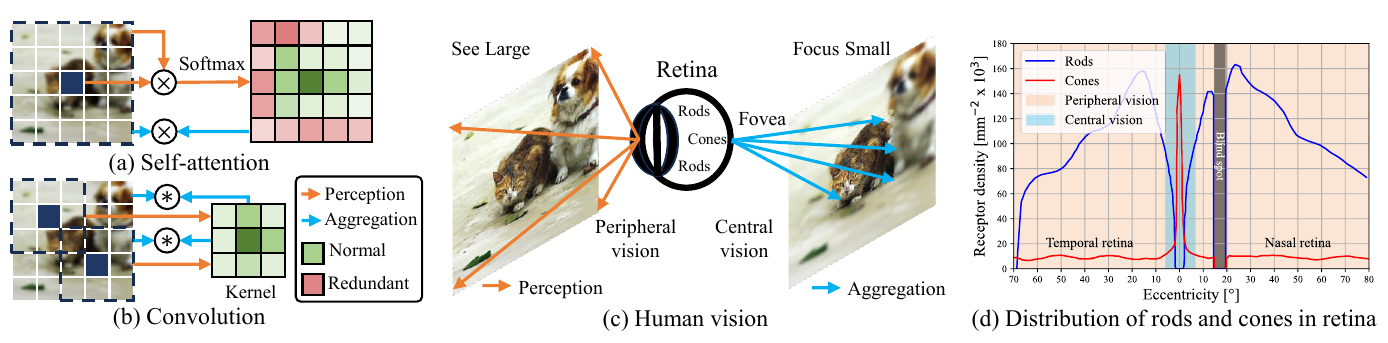}
  \caption{The mechanism of self attention (a) and convolution (b). (c) shows that the human vision system can ``See Large'' through the peripheral vision, and ``Focus Small'' through the central vision. (d) shows the distribution of rods and cones depending on the eccentricity from the fovea of the human eye. They contribute to the formation of extensive peripheral vision and focal central vision.}
  \label{fig:human-vision}
  \vspace{-15pt}
\end{figure*}

Essentially, contextual \textit{perception} and \textit{aggregation} are core processes for token mixing~\cite{tolstikhin2021mlp,fan2024lightweight,yang2022focal}, facilitating spatial information fusion. \textit{Perception} models contextual relationships among tokens, while \textit{aggregation} integrates token features based on corresponding relationships. In existing lightweight models, two dominant token mixing approaches, self-attention and convolution, employ distinct perception and aggregation processes. Specifically, self-attention employs global perception through holistic feature interaction and global aggregation via weighted sum of all features. Convolution uses the relative positional relationships among tokens for perception and aggregates features with static kernel weights. However, as shown in \cref{fig:human-vision}.(a) and (b), both approaches have limitations.
\begin{enumerate*}[label=(\arabic*)]
\item Self-attention often introduces excessive attention to regions lacking significant interconnections, leading to less critical aggregation, \eg, in less informative background~\cite{rao2021dynamicvit,liang2022not}. Besides, its perception and aggregation share the same mixing scope. The expansion of context in self-attention and its variants~\cite{liu2023efficientvit,fan2024lightweight,huang2022lightvit} comes at the expense of notable computational complexity. These hinder lightweight models from pursuing high representational ability under low computational budgets.
\item In convolution, the relationships among tokens modeled by the perception, \ie, the aggregation weights, are determined by the fixed kernel weights. Consequently, while efficient, convolution lacks sensitivity to varying contextual neighborhoods. This imposes constraints on the expressiveness of lightweight models, especially considering that the model capabilities of lightweight networks are inherently limited.
\end{enumerate*}
Given these, exploring a token mixing way for lightweight models with more effective and efficient perception and aggregation processes \textit{under limited computational costs} is imperative.

To this end, we first thoroughly inspect the intuitions underlying the processes of \textit{perception} and \textit{aggregation}. We discover that they align closely with the phenomenon of dynamic heteroscale vision ability in the efficient human vision system. Specifically, as shown in \cref{fig:human-vision}.(c), the human vision system follows dual-step mechanism: 
\begin{enumerate*}[label=(\arabic*)]
\item The broad overview of the scene is firstly captured through the \textit{peripheral vision's large-field perception}~\cite{Stewart,purves2001neurotransmission}, \ie, ``See Large''.
\item Subsequently, attention can be directed towards specific elements of the scene, enabling a detailed comprehension facilitated by the \textit{central vision's small-field aggregation}~\cite{Stewart,sterberg1935topography}, \ie, ``Focus Small''.
\end{enumerate*}
Such characteristic arises from distinct spatial distribution and vision abilities of two types of photoreceptor cells in the retina~\cite{purves2001neurotransmission,9261351}, \ie, rod cells and cone cells, as shown in \cref{fig:human-vision}.(d). Rod cells are widely distributed in the peripheral regions of retina~\cite{sterberg1935topography} and produce relatively unsharp images with limited spatial detail~\cite{wandell1995foundations}. However, they exhibit broad responses across the visible spectrum and contribute to large-field peripheral vision in conjunction with cone cells in the retina periphery~\cite{rodsp}, allowing ``See Large''. Furthermore, cone cells are primarily concentrated in the fovea, a small area for central vision~\cite{variationr}. The fovea contains a high density of cone cells, which constitute the sharpest region capable of capturing fine details and complex features~\cite{wandell1995foundations,Jonas,Tyagi_2018}, enabling ``Focus Small''. Guided by efficient large-field perception of the peripheral photoreceptor cells, the fovea can effectively focus on precise imaging of subtle features via small-field aggregation~\cite{purves2001neurotransmission}. This ``See Large, Focus Small'' approach empowers the human vision system to process visual information swiftly and proficiently~\cite{wandell1995foundations}, thereby facilitating accurate and efficient visual comprehension.

These inspections motivate us to design effective and efficient vision networks with the ability to perceive large fields and aggregate small fields. To this end, we first propose a novel operation, \textbf{L}arge-\textbf{S}mall (LS) convolution, which aims to emulate the ``See Large, Focus Small'' strategy observed in human vision system, thereby extracting discriminative visual patterns. Generally, LS convolution employs a large-kernel \textit{static} convolution for large-field perception and a small-kernel \textit{dynamic} convolution for small-field aggregation. Rather than simply combining large-kernel and small-kernel convolutions, it firstly leverages broad contextual information captured by large-kernel depth-wise convolution to model the spatial relationships. Then, parameterized by them, a small-kernel dynamic convolution operation with group mechanism is constructed to fuse features within highly related visual field. In this way, large-kernel static convolution well perceive the enlarged neighborhood information, leading to improved relationship modeling, like the peripheral vision system. Furthermore, benefiting from this, small-kernel dynamic convolution can adaptively aggregate the intricate visual features in small surroundings, enabling detailed visual understanding like the central vision system. Meanwhile, we delicately design LS convolution efficiently with depth-wise convolution and group mechanism. The aggregation scope is limited in a small region. These well ensure the low complexity of both perception and aggregation processes. Consequently, our LS convolution prioritizes both the performance and efficiency, enabling lightweight models to fully harness the representational capability under low computational costs.

We consider LS convolution as the fundamental operation of token mixing and integrate it with other common architecture designs to form a LS block. Building upon the LS block, we present a new family of lightweight models, dubbed LSNet. Extensive experiments demonstrate that LSNet achieves 
superior performance and efficiency compared with existing state-of-the-art lightweight models in various vision tasks~\cite{deng2009imagenet,lin2014microsoft,zhou2017scene}. We hope that LSNet can serve as a strong baseline and inspire further advancements in the field of lightweight and efficient models.

\section{Related Work}
\textbf{Efficient CNNs.} CNNs have emerged as the fundamental network architecture in various vision tasks~\cite{redmon2016you,long2015fully,bolya2019yolact,dong2015image,wang2023hierarchical,ding2023exploring} over the past decade. To facilitate their practical applications, researchers have devoted significant efforts to designing lightweight and efficient networks~\cite{howard2017mobilenets,howard2019searching,ding2021repvgg,ma2018shufflenet,wang2024repvit,tan2019efficientnet,ding2019acnet}. For example, MobileNet~\cite{howard2017mobilenets} and Xception~\cite{chollet2017xception} proposes architectures utilizing depth-wise separable convolutions. MobileNetV2~\cite{sandler2018mobilenetv2} introduces inverted residual blocks with linear bottleneck for improving efficiency. ShuffleNet~\cite{zhang2018shufflenet} and ShuffleNetV2~\cite{ma2018shufflenet} incorporate channel shuffling and channel split operations to enhance group information exchange. Hardware-aware neural architecture search (NAS) has also been explored to obtain compact vision networks~\cite{howard2019searching,tan2019efficientnet}.
Meanwhile, considering the limited receptive field, some works have explored enhancing lightweight CNNs' capability for modeling long-range dependencies~\cite{peng2017large,zhang2022parc,huang2023adaptive}. For example, ParC-Net~\cite{zhang2022parc} introduces position aware circular convolution to boast a global receptive field. AFFNet~\cite{huang2023adaptive} presents adaptive frequency filtering for global convolution via a circular padding.

\textbf{Efficient ViTs.} Later, since the inception of Vision Transformer~\cite{dosovitskiy2020image}, transformer-based architectures have gained significant popularity in the field of computer vision. ViTs have been adapted to diverse vision tasks and shown superior performance~\cite{zhang2022topformer,esser2021taming}. Meanwhile, efforts have been made to enhance the efficiency, resulting in lightweight ViTs for practical deployments~\cite{mehta2022separable,li2022efficientformer,vasu2023fastvit,wang2023cait}. For example, MobileViT~\cite{mehta2021mobilevit} combines MobileNet blocks and MHSA blocks, achieving a hybrid architecture. EdgeViT~\cite{pan2022edgevits} proposes the integration of self-attention and convolutions to achieve cost-effective information exchange. Besides, to alleviate the inference bottleneck, EfficientFormer~\cite{li2022efficientformer} presents a dimension-consistent design paradigm that enhances the latency and performance trade-off. FastViT~\cite{vasu2023fastvit} introduces structural re-parameterization and large-kernel convolutions to enhance hybrid ViTs.

\textbf{Efficient Token Mixing.} CNNs and ViTs adopt different token mixing ways, \ie, convolution and self-attention, respectively, along with distinct perception and aggregation processes. Based on them, to develop lightweight vision networks, researchers have explored different efficient token mixing ways for spatial information exchange. For example, for convolution, Involution~\cite{li2021involution} leverages MLP for perception to derive the aggregation weights conditioned on single pixel. CondConv~\cite{yang2019condconv} proposes per-example routing with global context to linearly combining multiple convolution kernels. For self-attention, EdgeNeXt~\cite{maaz2022edgenext} presents split depth-wise transpose attention (SDTA) to mix multi-scale features. PVTv2~\cite{wang2022pvt} employs linear spatial reduction attention (LSRA) to achieve linear computational complexity for the attention layer. EfficientViT~\cite{liu2023efficientvit} designs the cascaded group attention to enhance capability efficiently. 

\section{Methodology}

\begin{figure}[t]
  \centering
  \includegraphics[width=1.0\columnwidth]{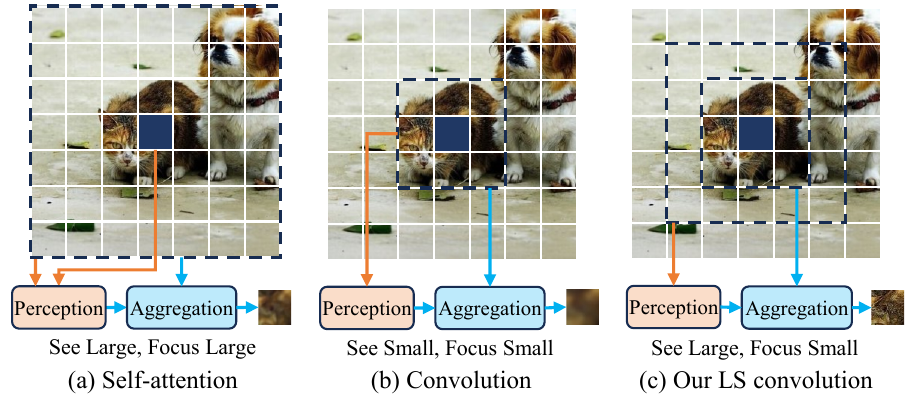}
  \caption{Comparison of self-attention, convolution, and LS conv.}
  \label{fig:percetion-aggregation}
  \vspace{-10pt}
\end{figure}

\subsection{Revisiting Self-Attention and Convolution}
Self-attention and convolution are two prominent token mixing ways~\cite{yu2022metaformer} for modeling visual features in existing lightweight networks. 
For an input image, given its feature map $X\in \mathbb{R}^{H\times W\times C}$ where $H\times W$ is the spatial resolution and $C$ is the number of channels, token mixing generates the feature representation $y_i \in \mathbb{R}^C$ for each token $x_i \in \mathbb{R}^C$ based on its contextual region $\mathcal{N}(x_i)$ by:
\begin{equation}
  \small
  \setlength{\abovedisplayskip}{5pt}
  \setlength{\belowdisplayskip}{5pt}
  y_i = \mathcal{A}(\mathcal{P}(x_i, \mathcal{N}(x_i)), \mathcal{N}(x_i))
  \label{eq:mixing},
\end{equation}
where $\mathcal{P}$ denotes perception, involving extracting contextual information and capturing the relationships among tokens, and $\mathcal{A}$ denotes aggregation, integrating the features based on the outcome of perception and enabling the incorporation of information from other tokens.

In \textsf{self-attention}, its perception $\mathcal{P}_{attn}$ obtains the attention scores between $x_i$ and $X$ through the pairwise correlations after \textsf{softmax} normalization. Its aggregation $\mathcal{A}_{attn}$ weights the features of $X$ by attention scores to obtain $y_i$. As shown in \cref{fig:percetion-aggregation}.(a), the process can be summarized as:
\begin{small}
  \setlength{\abovedisplayskip}{-5pt}
  \setlength{\belowdisplayskip}{5pt}
  \begin{gather}
  y_i = \mathcal{A}_{attn}(\mathcal{P}_{attn}(x_i, X), X) = \mathcal{P}_{attn}(x_i, X) (XW_v); \\
  \mathcal{P}_{attn}(x_i, X) = \text{softmax}((x_iW_q)(XW_k)^T),
  \label{eq:attention}
\end{gather}
\end{small}where $W_q$, $W_k$ and $W_v$ are the projection matrices. It can be observed that $\mathcal{P}_{attn}$ and $\mathcal{A}_{attn}$ involve redundant attention and excessive aggregation in less informative regions~\cite{rao2021dynamicvit,liang2022not}, limiting the efficacy of lightweight models. Moreover, they operate at the same contextual scale for $x_i$. Such a homoscale property leads to the notable computational complexity when increasing the mixing scope $\mathcal{N}(x_i)$, imposing challenges in expanding the perception context under low computational budgets. Thus, self-attention and its variants in existing lightweight models~\cite{liu2023efficientvit,fan2024lightweight} struggle to achieve an optimal balance between representation capability and efficiency with limited computation cost~\cite{huang2023adaptive}.

For \textsf{convolution} with the kernel size of $K$, the contextual region is the neighborhood of size $K\times K$ centered around $x_i$, denoted as $\mathcal{N}_{K}(x_i)$. The perception $\mathcal{P}_{conv}$ utilizes the relative positions between $x_i$ and $\mathcal{N}_{K}(x_i)$ to derive the aggregation weights. For each $x_j \in \mathcal{N}_{K}(x_i)$, its aggregation weight is the value at the corresponding relative position in the fixed convolutional kernel weights $W_{conv}$. The aggregation $\mathcal{A}_{conv}$ then leverages the weights to convolve the features in $\mathcal{N}_{K}(x_i)$. As shown in \cref{fig:percetion-aggregation}.(b), the whole process can be formulated as:
\begin{equation}
  \small
  \setlength{\abovedisplayskip}{5pt}
  \setlength{\belowdisplayskip}{5pt}
  \begin{split}
  y_i &= \mathcal{A}_{conv}(\mathcal{P}_{conv}(x_i, \mathcal{N}_{K}(x_i)), \mathcal{N}_{K}(x_i)) \\ &= \mathcal{P}_{conv}(x_i, \mathcal{N}_{K}(x_i)) \circledast \mathcal{N}_{K}(x_i); \\
  \end{split}
\end{equation}
\vspace{-10pt}
\begin{small}
\setlength{\abovedisplayskip}{5pt}
\setlength{\belowdisplayskip}{5pt}
\begin{gather}
  \mathcal{P}_{conv}(x_i, \mathcal{N}_{K}(x_i)) = W_{conv},
\end{gather}
\end{small}
where $\circledast$ denotes the convolution operation. It can be observed that the token mixing scope in convolution is determined by kernel size $K$ which is usually small for lightweight models, thus resulting in a limited perception range. Besides, the relationships among tokens modeled by the perception $\mathcal{P}_{conv}$, \ie, the aggregation weights, depend only on the relative positions and thus are shared and fixed for all tokens. It prevents tokens from adapting to their related context, restricting the expressive ability. Such limitation becomes particularly pronounced considering the inherently small modeling capability of lightweight networks.

\subsection{LS (\underline{L}arge-\underline{S}mall) Convolution}
Inspired by dynamic heteroscale vision ability exhibited by human vision system~\cite{sterberg1935topography,purves2001neurotransmission,rodsp}, we introduce a novel ``See Large, Focus Small'' strategy for the perception and aggregation processes, aiming for efficient and effective token mixing in lightweight models, as shown in \cref{fig:percetion-aggregation}.(c). Our approach enables the effective collection of comprehensive contextual information and modeling of the relationships by large-field perception. It further facilitates detailed visual representations through efficient fusion in highly related surroundings by small-field aggregation. Specifically, for token $x_i$, with the contextual regions of perception and aggregation as $\mathcal{N}_P(x_i)$ and $\mathcal{N}_A(x_i)$, respectively, where $\mathcal{N}_P(x_i)$ encompasses a larger spatial extent compared with $\mathcal{N}_A(x_i)$, the process can be formulated as:
\begin{equation}
  \small
  \setlength{\abovedisplayskip}{5pt}
  \setlength{\belowdisplayskip}{5pt}
  y_i = \mathcal{A}(\mathcal{P}(x_i, \mathcal{N}_P(x_i)), \mathcal{N}_A(x_i))
  \label{eq:ls}.
\end{equation}
It can be observed that 
\begin{enumerate*}[label=(\arabic*)]
\item The perception $\mathcal{P}$ and aggregation $\mathcal{A}$ involves different contextual scopes, \ie, $\mathcal{N}_P(x_i)$ and $\mathcal{N}_A(x_i)$, respectively, allowing for utilizing heteroscale contextual information and capturing both the overall context and fine-grained details.
\item For the perception with a large spatial extent, cost-effective operations, such as large-kernel depth-wise convolution, can be employed. The perception context can thus be enlarged with minimal overhead.
\item For the aggregation with a small surrounding region, we can adopt adaptive weighted feature summation. Due to the limited range of aggregation, the efficiency can be guaranteed with low computation cost and the less important aggregation in self-attention can be mitigated.
\end{enumerate*}

\begin{figure*}[t]
  \centering
  \includegraphics[width=0.85\textwidth]{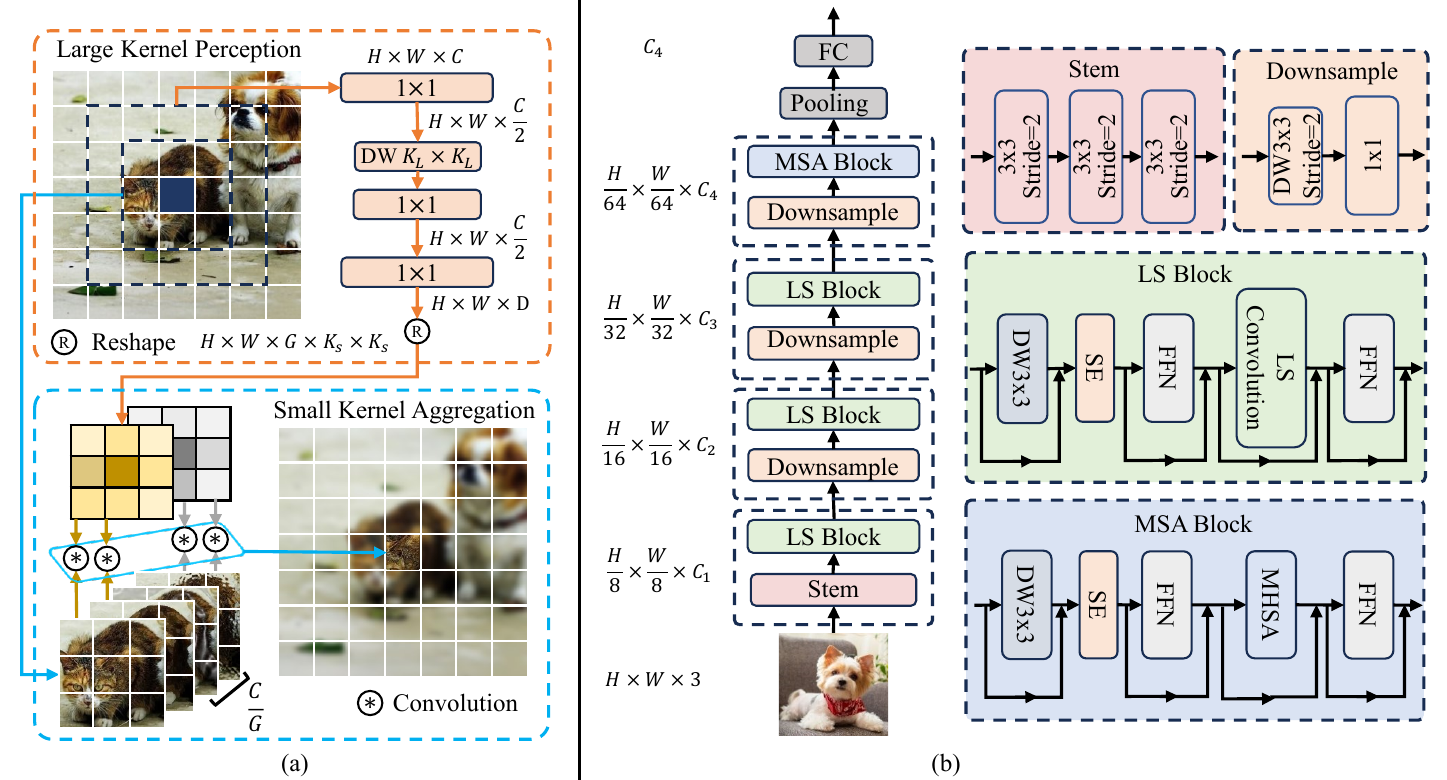}
  \caption{(a) The illustration of our proposed LS convolution. (b) The illustration of our proposed LSNet. LSNet has four stages with $\frac{H}{8}\times \frac{W}{8}$, $\frac{H}{16}\times \frac{W}{16}$, $\frac{H}{32}\times \frac{W}{32}$, and $\frac{H}{64}\times \frac{W}{64}$ resolutions respectively, where $H$ and $W$ denote the width and height of the input image. $C$ represents the channel dimension. The norm layer and nonlinearity are omitted for simplicity.}
  \label{fig:lsnet}
  \vspace{-10pt}
\end{figure*}

Based on these, we present a novel LS (\textbf{L}arge-\textbf{S}mall) convolution. As shown in \cref{fig:lsnet}.(a), for each token, it introduces two steps: \begin{enumerate*}[label=(\arabic*)]
\item Large-kernel perception $\mathcal{P}_{ls}$ models the neighborhood relationships with the enlarged receptive field through large-kernel static convolutions.
\item Small-kernel aggregation $\mathcal{A}_{ls}$ adaptively integrates the surrounding features through small-kernel dynamic convolution.
\end{enumerate*}

\textbf{Large-Kernel Perception} (\textbf{LKP}) adopts the design of a large-kernel bottleneck block. Given visual feature map $X\in \mathbb{R}^{H\times W\times C}$, we initially utilize the point-wise convolution (PW) to project the tokens into a lower channel dimension, \ie, $\frac{C}{2}$ by default, to reduce the computational cost and make the model lightweight as possible. For $x_i$, we then employ large-kernel depth-wise convolution (DW) with the kernel size of $K_L\times K_L$ to efficiently capture large-field spatial contextual information of $\mathcal{N}_{K_L}(x_i)$, where $\mathcal{N}_{K_L}(x_i)$ denotes the surroundings of size $K_L\times K_L$ centered around $x_i$. The large-kernel DW can well expand the receptive field and enhance the context perception capability under minimal cost. We then leverage point-wise convolutions (PW) to model the spatial relationships among tokens, \ie, generating the context-adaptive weights $W\in \mathbb{R}^{H\times W\times D}$ for the aggregation step. The whole process can be formulated as:
\begin{equation}
  \small
  \setlength{\abovedisplayskip}{5pt}
  \setlength{\belowdisplayskip}{5pt}
  \begin{split}
    w_i & = \mathcal{P}_{ls}(x_i, \mathcal{N}_{K_L}(x_i)) \\
    & = \text{PW}(\text{DW}_{K_L \times K_L}(\text{PW}(\mathcal{N}_{K_L}(x_i)))),
  \end{split}
\end{equation}
where $w_i \in \mathbb{R}^{D}$ is the generated weights for $x_i$.

\textbf{Small-Kernel Aggregation} (\textbf{SKA}) employs the design of grouped dynamic convolutions. For the visual feature map $X\in \mathbb{R}^{H\times W \times C}$, we divide its channels into $G$ groups. Each group containing $\frac{C}{G}$ channels and the channels in the same group share the aggregation weights, to reduce the memory overhead and computational cost for lightweight models. For each $x_i$, we reshape its corresponding weights $w_i \in \mathbb{R}^{D}$ generated by large-kernel perception to obtain $w_i^* \in \mathbb{R}^{G\times K_S \times K_S}$, where $K_S \times K_S$ is the small kernel size. We then leverage $w_i^*$ to aggregate its highly related context of $\mathcal{N}_{K_S}(x_i)$, where $\mathcal{N}_{K_S}(x_i)$ represents the neighborhood of size $K_S\times K_S$ centered around $x_i$. Specifically, we denote the $c$-th channel of $x_i$ as $x_{ic}$, which belongs to the $g$-th channel group. We obtain its aggregated feature representation $y_{ic}$ through the convolution operation between $\mathcal{N}_{K_S}(x_{ic})$ and $w_{ig}^* \in \mathbb{R}^{K_S\times K_S}$. In this way, the adaptive fine-grained features can be effectively represented, making model sensitive to dynamic and complex changes in diverse contexts. The whole process can be formulated as:
\begin{equation}
  \small
  \setlength{\abovedisplayskip}{5pt}
  \setlength{\belowdisplayskip}{5pt}
  y_{ic} = \mathcal{A}_{ls}(w_{ig}^*, \mathcal{N}_{K_S}(x_{ic}))=w_{ig}^*\circledast \mathcal{N}_{K_S}(x_{ic})
  \label{eq:ska}.
\end{equation}

In contrast to simply combining large-kernel with small-kernel conv, and other dynamic convs, our LKP utilizes enriched large-field visual perception to guide adaptive feature fusion within highly related context by SKA. This enables more discriminative representations for intricate visual information. Thus, LS conv shows superiority over them, as shown in \cref{tab:dynamic} and \cref{tab:lkp-ska}. We also present the comparisons from mathematical perspectives in supplementary.

\textbf{Complexity Analysis.} The computation of LS convolution mainly consists of three parts: point-wise convolutions in $\mathcal{P}_{ls}$, depth-wise convolution with kernel size of $K_L$ in $P_{ls}$, and convolution aggregation with kernel size of $K_S$ in $\mathcal{A}_{ls}$. Their corresponding computations are $\mathcal{O}(\frac{3HWC^2}{4}+\frac{HWCD}{2})$, $\mathcal{O}(\frac{HWCK_L^2}{2})$, and $\mathcal{O}(HWCK_S^2)$, respectively. Therefore, the total amount is $\mathcal{O}(\frac{HWC}{4}(3C+2K_L^2 + (2G+4)K_S^2))$, enjoying the linear computational complexity with respect to the input resolution.

\subsection{LSNet: \underline{L}arge-\underline{S}mall Network}
Using LS convolution as the primary operation, we present the basic block, \ie, LS block, and the lightweight model design, \ie, LSNet, as shown in \cref{fig:lsnet}.(b).

LS Block leverages LS convolution to perform effective token mixing. Skip connection is adopted to facilitate model optimization. Besides, we utilize the extra depth-wise convolution and SE layer~\cite{hu2018squeeze} to enhance model capability by introducing more local inductive bias~\cite{dai2021coatnet,liu2023efficientvit}. Feed forward network (FFN) is adopted for channel mixing.

LSNet utilizes overlapping patch embedding~\cite{xiao2021early} to project the input image into the visual feature map. For downsampling, we leverage the depth-wise and point-wise convolution to reduce the spatial resolution and modulate the channel dimension, respectively. Besides, we stack LS blocks in the top three stages. In the last stage, we adopt the MSA block to capture long-range dependencies due to the small resolution, following~\cite{vasu2023fastvit,mehta2021mobilevit}. MSA block incorporates multi-head self-attention (MHSA), and we utilize the same depth-wise convolution and SE layer to introduce more local structural information like LS block. 

We build three LSNet variants for different computational budgets. The LSNet with tiny size (LSNet-T), small size (LSNet-S), and base size (LSNet-B) has 0.3G, 0.5G, and 1.3G FLOPs, respectively. Following~\cite{graham2021levit,liu2023efficientvit}, we employ more blocks in late stages, due to that processing on early stages with higher resolution is more time consuming. We empirically use $K_L=7$, $K_S=3$, and $G=\frac{C}{8}$ for all model variants by default, following~\cite{liu2022convnet,ding2021repvgg}. The architectural details can be found in the supplementary.

\begin{table}
  \caption{\textbf{Classification results on ImageNet-1K.} The throughput is tested on a Nvidia RTX3090 with maximum power-of-two batch size that fits in memory, following~\cite{liu2023efficientvit,huang2023adaptive}. * denotes the results with distillation using the RegNetY-16GF~\cite{radosavovic2020designing} with 82.9\% top-1 accuracy as the teacher model. EFormer denotes EfficientFormer.}
  \label{tab:imagenet}
  \setlength{\tabcolsep}{4pt}
  \small
  \centering
  \begin{tabular}{lccccc}
  \toprule
  Model & \makecell{Params\\(M)} & \makecell{FLOPs\\(G)} & \makecell{Throughput\\(img/s)} & \makecell{Top-1\\(\%)} \\
  \toprule
  EdgeNeXt-XXS~\cite{maaz2022edgenext} & 1.3 & 0.3 & 5089 & 71.2  \\
  FasterNet-T0~\cite{chen2023run} & 3.9 & 0.3 & 14467 & 71.9 \\
  ShuffleNetV2~\cite{ma2018shufflenet} & 3.5 & 0.3 & 9593 & 72.6 \\
  AFFNet-ET~\cite{huang2023adaptive} & 1.4 & 0.4 & 2877 & 73.0  \\
  EfficientViT-M3~\cite{liu2023efficientvit} & 6.9 & 0.3 & 14613 & 73.4  \\
  StarNet-S1~\cite{ma2024rewrite} & 2.9 & 0.4 & 5034 & 73.5 \\
  \rowcolor[gray]{0.92}
  \textbf{LSNet-T} & 11.4 & 0.3  & 14708 & \textbf{74.9} \\
  \rowcolor[gray]{0.92}
  \textbf{LSNet-T}* & 11.4 & 0.3  & 14708 & \textbf{76.1} \\
  \midrule
  EdgeNeXt-XS~\cite{maaz2022edgenext} & 2.3 & 0.5 & 3118 & 75.0\\
  PVT-Tiny~\cite{wang2021pyramid} & 13.2 & 1.9 & 2125 & 75.1\\
  MobileNetV3-L~\cite{howard2019searching} & 5.4 & 0.2 & 7921 & 75.2 \\
  FastViT-T8~\cite{vasu2023fastvit} & 3.6 & 0.7 & 3909 & 75.6 \\
  EFormerV2-S0*~\cite{li2023rethinking} & 3.5 & 0.4 & 1329 & 75.7 \\
  FasterNet-T1~\cite{chen2023run} & 7.6 & 0.9 & 8660 & 76.2 \\
  UniRepLKNet-A~\cite{ding2024unireplknet} & 4.4 & 0.6 & 3931 & 77.0 \\
  EfficientNet-B0~\cite{tan2019efficientnet} & 5.3 & 0.4 & 4481 & 77.1 \\
  PoolFormer-S12~\cite{yu2022metaformer} & 12.0 & 1.8 & 2769 & 77.2 \\
  SHViT-S3~\cite{yun2024shvit} & 14.2 & 0.6 & 8993 & 77.4 \\
  RepViT-M0.9~\cite{wang2024repvit} & 5.1 & 0.8 & 4817 & 77.4 \\
  \rowcolor[gray]{0.92}
  \textbf{LSNet-S} & 16.1 & 0.5 & 9023 & \textbf{77.8}  \\
  \rowcolor[gray]{0.92}
  \textbf{LSNet-S}* & 16.1 & 0.5 & 9023 & \textbf{79.0}  \\
  \midrule
  EdgeViT-XS~\cite{pan2022edgevits} & 6.7 & 1.1 & 2751 & 77.5 \\
  SwiftFormer-S*~\cite{shaker2023swiftformer} & 6.1 & 1.0 & 3376 & 78.5 \\
  UniRepLKNet-F~\cite{ding2024unireplknet} & 6.2 & 0.9 & 3209 & 78.6 \\
  FastViT-T12~\cite{vasu2023fastvit} & 6.8 & 1.4 & 2586 & 79.1 \\
  EFormer-L1*~\cite{li2022efficientformer} & 12.3 & 1.3 & 3280 & 79.2  \\
  EdgeNeXt-S~\cite{maaz2022edgenext} & 5.6 & 1.3 & 2128 & 79.4 \\
  RepViT-M1.1~\cite{wang2024repvit} & 8.2 & 1.3 & 3604 & 79.4 \\
  PVT-Small~\cite{wang2021pyramid} & 24.5 & 3.8 & 1160 & 79.8 \\
  AFFNet~\cite{huang2023adaptive} & 5.5 & 1.5 & 1355 & 79.8 \\
  \rowcolor[gray]{0.92}
  \textbf{LSNet-B} & 23.2 & 1.3 & 3996 & \textbf{80.3} \\
  \rowcolor[gray]{0.92}
  \textbf{LSNet-B}* & 23.2 & 1.3 & 3996 & \textbf{81.6} \\
  \bottomrule
  \end{tabular}
  \vspace{-10pt}
\end{table}

\section{Experiments}

\subsection{Image Classification}
We conduct experiments on ImageNet-1K~\cite{deng2009imagenet} under the same training recipe as~\cite{liu2023efficientvit,huang2023adaptive,pan2022edgevits} to assess the performance of LSNet on the image classification task. 

As shown in \cref{tab:imagenet}, we note that LSNet consistently achieves state-of-the-art performance across various computational costs. Besides, it shows the best trade-offs between accuracy and inference speed. For example, our LSNet-B outperforms the advanced AFFNet by 0.5\% top-1 accuracy with a nearly 3$\times$ faster inference speed. It also surpasses RepViT-M1.1 and FastViT-T12 with 0.9\% and 1.2\% top-1 accuracies with higher efficiency, respectively. For smaller models, our LSNet also obtains superior performance with lower computation costs. Specifically, LSNet-S outperforms UniRepLKNet-A and FasterNet-T1 significantly by 0.8\% and 1.6\% top-1 accuracies, respectively, along with higher throughput. Compared with StarNet-S1 and EfficientViT-M3, LSNet-T also improves the top-1 accuracy by 1.4\% and 1.5\%, respectively. These results well show the effectiveness and efficiency of our LSNet models.

\begin{table*}[t]
\caption{\textbf{Object detection and instance segmentation results on COCO.} AP$^{\rm b}$ and AP$^{\rm m}$ indicate bounding box AP and mask AP, respectively. Following common convention~\cite{vasu2023fastvit}, FLOPs (G) of backbone is measured on image crops of 512$\times$512.
}
\label{tab:coco} 
\centering
\small
\resizebox{\textwidth}{!}{
\begin{tabular}{l|c|ccc|ccc|ccc|ccc}
  \toprule
  \multirow{2}{*}{Backbone} & \multirow{2}{*}{FLOPs} &\multicolumn{6}{c|}{RetinaNet} &\multicolumn{6}{c}{Mask R-CNN} \\ 
  \cmidrule{3-14}
  &  & AP & AP$_{50}$ & AP$_{75}$ & AP$_S$ & AP$_M$ & AP$_L$ & AP$^{\rm b}$ &AP$_{50}^{\rm b}$ &AP$_{75}^{\rm b}$  &AP$^{\rm m}$ &AP$_{50}^{\rm m}$ &AP$_{75}^{\rm m}$ \\
  \toprule 
  MobileNetV2~\cite{sandler2018mobilenetv2} & 1.6 & 28.3 & 46.7 & 29.3 & 14.8 & 30.7 & 38.1 & 29.6 & 48.3 & 31.5 & 27.2 & 45.2 & 28.6 \\
  MobileNetV3~\cite{howard2019searching} & 1.1 & 29.9 & 49.3 & 30.8 & 14.9 & 33.3 & 41.1 & 29.2 & 48.6 & 30.3 & 27.1 & 45.5 & 28.2 \\
  FairNAS-C~\cite{chu2021fairnas} & 1.7 & 31.2 & 50.8 & 32.7 & 16.3 & 34.4 & 42.3 & 31.8 & 51.2 & 33.8 & 29.4 & 48.3 & 31.0  \\
  EfficientViT-M4~\cite{liu2023efficientvit} & 1.6 & 32.7 & 52.2 & 34.1 & 17.6 & 35.3 & 46.0 & 32.8 & 54.4 & 34.5 & 31.0 & 51.2 & 32.2 \\
  StarNet-S1~\cite{ma2024rewrite} & 2.2 & 33.6 & 53.3 & 35.1 & \textbf{18.3} & 36.0 & 47.0 & 33.8 & 56.1 & 35.5 & 31.9 & 52.9 & 33.4 \\
  \rowcolor[gray]{0.92}
  \textbf{LSNet-T} & 1.5 & \textbf{34.2} & \textbf{54.6} & \textbf{35.2} & 17.8 & \textbf{37.1} & \textbf{48.5} & \textbf{35.0} & \textbf{57.0} & \textbf{37.3} & \textbf{32.7} & \textbf{53.8} & \textbf{34.3} \\
  \midrule
  ResNet18~\cite{he2016deep} & 9.5 & 31.8 & 49.6 & 33.6 & 16.3 & 34.3 & 43.2 & 34.0 & 54.0 & 36.7 & 31.2 & 51.0 & 32.7\\ 
  DFvT-T~\cite{gao2022doubly} & 6.9 & - & - & - & - & - & - & 34.8 & 56.9 & 37.0 & 32.6 & 53.7 & 34.5 \\
  EfficientViT-M5~\cite{liu2023efficientvit} & 2.8 & 34.3 & 54.2 & 36.1 & 18.0 & 36.9 & 48.2 & 34.9 & 57.0 & 37.0 & 32.8 & 53.7 & 34.6 \\
  SHViT-S3~\cite{yun2024shvit} & 3.0 & 36.1 & 56.6 & 38.0 & 19.9 & 39.1 & 50.8 & 36.9 & 59.4 & 39.6 & 34.4 & 56.3 & 36.1 \\
  \rowcolor[gray]{0.92}
  \textbf{LSNet-S} & 2.6 & \textbf{36.7} & \textbf{57.2} & \textbf{38.6} & \textbf{20.0} & \textbf{39.7} & \textbf{51.8} & \textbf{37.4} & \textbf{59.9} & \textbf{39.8} & \textbf{34.8} & \textbf{56.8} & \textbf{36.6} \\
  \midrule 
  ResNet50~\cite{he2016deep} & 21.4 & 36.3 & 55.3 & 38.6 & 19.3 & 40.0 & 48.8 & 38.0 & 58.6 & 41.4 & 34.4 & 55.1 & 36.7\\
  PVT-Tiny~\cite{wang2021pyramid} & 11.8 & 36.7 & 56.9 & 38.9 & \textbf{22.6} & 38.8 & 50.0  & 36.7 & 59.2 & 39.3 & 35.1 & 56.7 & 37.3 \\
  PoolFormer-S12~\cite{yu2022metaformer} & 9.5 & 36.2 & 56.2 & 38.2 & 20.8 & 39.1 & 48.0 & 37.3 & 59.0 & 40.1 & 34.6 & 55.8 & 36.9\\
  FasterNet-S~\cite{chen2023run} & 23.8 & - & - & - & - &- & - & 39.9 & 61.2 & 43.6 & 36.9 & 58.1 & 39.7  \\
  FastViT-SA12~\cite{vasu2023fastvit} & 7.7 & - & - & - & - & - & - &38.9 & 60.5 & 42.2 & 35.9 & 57.6 & 38.1 \\
  RepViT-M1.1~\cite{wang2024repvit} & 7.0 & - & - & - & - &- & - & 39.8 &  61.9 & 43.5 & 37.2 & 58.8 & \textbf{40.1} \\
  \rowcolor[gray]{0.92}
  \textbf{LSNet-B} & 6.2 & \textbf{39.2} & \textbf{60.0} & \textbf{41.5} & 22.1 & \textbf{43.0} & \textbf{52.9} & \textbf{40.8} & \textbf{63.4} & \textbf{44.0} & \textbf{37.8} & \textbf{60.5} & \textbf{40.1} \\
  \bottomrule
\end{tabular}
}
\vspace{-10pt}
\end{table*}

\subsection{Downstream Tasks}
\textbf{Object Detection and Instance Segmentation.} We evaluate LSNet on object detection and instance segmentation tasks to verify its transferability. Following~\cite{pan2022edgevits,liu2023efficientvit}, we integrate LSNet into RetinaNet~\cite{lin2017focal} and Mask R-CNN~\cite{he2017mask} and conduct experiments on COCO-2017~\cite{lin2014microsoft}. As shown in \cref{tab:coco}, our LSNet consistently shows superior performance compared with competitor models. Specifically, in the RetinaNet framework for object detection, LSNet-T outperforms StarNet-S1 by 0.6 AP and 1.3 AP$_{50}$ under notably less computational cost. For large models, our LSNet-B also surpasses PoolFormer-S12 and PVT-Tiny with considerable margins of 3.0 AP and 2.5 AP, respectively. When integrated into the Mask R-CNN framework for object detection and instance segmentation, LSNet-S obtains the favorable improvements of 0.5 AP$^b$ and 2.5 AP$^b$ over SHViT-S3 and EfficientViT-M5, respectively. Compared with RepViT-M1.1, LSNet-B also achieves 1.0 higher AP$^{\rm b}$ and 0.6 higher AP$^{\rm m}$, demonstrating the superiority in transferring.

\textbf{Semantic Segmentation.} We evaluate LSNet on the semantic segmentation task by conducting experiments on ADE20K~\cite{zhou2017scene}. Following~\cite{pan2022edgevits,li2022efficientformer}, we incorporate LSNet in the Semantic FPN~\cite{kirillov2019panoptic} segmentation model. As shown in \cref{tab:ade20k}, LSNet performs clearly better in all comparisons across different model scales. It can achieve superior performance under low computational costs. Specifically, LSNet-T significantly outperforms VAN-B0 by 1.6 mIoU, and it also achieves 2.9 higher mIoU over PVTv2-B0. For larger models, LSNet-S obtains the improvements of 0.4 mIoU and 1.0 mIoU over the advanced RepViT-M1.1 and SHViT-S3, respectively, with lower computational complexity. Additionally, LSNet-B surpasses SwiftFormer-L1 and FastViT-SA24 by margins of 1.6 and 2.0 mIoUs respectively. These results further show the efficacy of LSNet.

\begin{table}[t]
  \caption{\textbf{Semantic segmentation on ADE20K.} Following~\cite{vasu2023fastvit}, FLOPs (G) of backbone are measured on image crops of 512$\times$512.}
  \label{tab:ade20k}
  \small
  \subfloat{
    \hspace{-10pt}
    \resizebox{0.48\columnwidth}{!}{
    \setlength{\tabcolsep}{3pt}
    \begin{tabular}{l|c|c}
    \toprule
    Backbone & FLOPs & mIoU\\
    \toprule
    StarNet-S1 & 2.2 & 36.0 \\
    MobileNetV3 & 1.1 & 37.0 \\
    PVTv2-B0 & 3.8 & 37.2 \\
    VAN-B0 & 4.5 & 38.5 \\
    \rowcolor[gray]{0.92}
    \textbf{LSNet-T} & 1.5 & \textbf{40.1} \\
    \midrule
    EdgeViT-XXS & 3.2 & 39.7 \\
    SHViT-S3 & 3.0 & 40.0 \\
    FastViT-SA12 & 7.7 & 38.0 \\
    RepViT-M1.1 & 7.0 & 40.6 \\
    \rowcolor[gray]{0.92}
    \textbf{LSNet-S} & 2.6 & \textbf{41.0} \\
    \bottomrule
    \end{tabular} 
    }
  }
  \subfloat{
    \resizebox{0.525\columnwidth}{!}{
    \setlength{\tabcolsep}{3pt}
    \begin{tabular}{l|c|c}
    \toprule
    Backbone & FLOPs & mIoU\\
    \toprule
    EFormer-L1 & 6.8 & 38.9 \\
    PVT-Small & 23.1 & 39.8 \\
    PoolFormer-S24 & 17.8 & 40.3 \\
    FastViT-SA24 & 15.0 & 41.0 \\
    EdgeViT-XS & 6.3 & 41.4 \\
    SwiftFormer-L1 & 8.3 & 41.4 \\
    Swin-T & 25.6 & 41.5 \\
    EFormerV2-S2 & 7.3 & 42.4 \\
    PVTv2-B1 & 12.8 & 42.5 \\
    \rowcolor[gray]{0.92}
    \textbf{LSNet-B} & 6.2 & \textbf{43.0} \\
    \bottomrule
    \end{tabular} 
    }
  }
  \vspace{-14pt}
\end{table}

\subsection{Robustness Evaluation}
We conduct robustness evaluation for LSNet on various benchmarks, including ImageNet-C~\cite{hendrycks2019benchmarking}, ImageNet-A~\cite{hendrycks2021natural}, ImageNet-R~\cite{hendrycks2021many}, and ImageNet-Sketch~\cite{wang2019learning}. Following~\cite{vasu2023fastvit,mao2022towards,liu2022convnet}, we report mean corruption error (lower is better) for ImageNet-C and top-1 accuracies for other datasets. As shown in \Cref{tab:robust}, LSNet shows strong domain generalization capabilities and promising robustness to corruptions, achieving state-of-the-art performance. For example, compared with UniRepLKNet-A, LSNet-B exhibits a 1.3 mCE reduction on ImageNet-C, along with top-1 accuracy gains of 1.2\%, 1.5\%, and 1.5\% on ImageNet-A, ImageNet-R, and ImageNet-Sketch, respectively. LSNet-T also outperforms StarNet-S1 significantly by 2.2\% and 3.7\% on ImageNet-A and ImageNet-Sketch, respectively, highlighting the robust generalization ability.

\begin{table}[t]
\caption{\textbf{Robustness evaluation results on benchmark datasets}, where we report mCE for ImageNet-C and top-1 accuracies for ImageNet-A, ImageNet-R, and ImageNet-Sketch.}
\label{tab:robust}
\small
\centering
  \resizebox{1.0\columnwidth}{!}{
  \begin{tabular}{lccccc}
  \toprule
  Model & FLOPs & C ($\downarrow$) & A & R & SK               \\
  \toprule
  FasterNet-T0~\cite{chen2023run} & 0.3 & 89.8 & 2.3 & 28.6 & 16.3 \\
  EdgeNeXt-XXS~\cite{maaz2022edgenext} & 0.3 & 94.6 & 3.6 & 29.5 & 18.5 \\
  EfficientViT-M3~\cite{liu2023efficientvit} & 0.3 & 71.1 & 5.2 & 36.1 & 23.4 \\
  StarNet-S1~\cite{ma2024rewrite} & 0.4 & 77.5 & 4.5 & 34.1 & 21.8 \\
  \rowcolor[gray]{0.92}
  \textbf{LSNet-T} & 0.3 & \textbf{68.2} & \textbf{6.7} & \textbf{38.5} & \textbf{25.5} \\
  \midrule
  FastViT-T8~\cite{vasu2023fastvit} & 0.7 & 72.1 & 6.9 & 36.8 & 25.5 \\
  PVTv2-B0~\cite{wang2022pvt} & 0.6 & 75.4 & 4.2 & 34.2 & 21.5  \\
  EdgeNeXt-XS~\cite{maaz2022edgenext} & 0.5 & 88.4 & 6.3 & 32.5 & 22.0 \\
  UniRepLKNet-A~\cite{ding2024unireplknet} & 0.6 & 67.0 & 8.4 & 37.9 & 26.0 \\
  \rowcolor[gray]{0.92}
  \textbf{LSNet-S} & 0.5 & \textbf{65.7} & \textbf{9.6} & \textbf{39.4} & \textbf{27.5} \\
  \midrule
  PVT-Tiny~\cite{wang2021pyramid} & 1.9 & 79.6 & 7.9 & 33.9 & 21.5 \\
  PoolFormer-S12~\cite{yu2022metaformer} & 1.8 & 67.7 & 6.9 & 37.7 & 25.2 \\
  FasterNet-T2~\cite{chen2023run} & 1.9 & 70.8 & 8.7 & 40.5 & 27.2 \\
  EdgeNeXt-S~\cite{maaz2022edgenext} & 1.3 & 72.1 & 11.9 & 40.1 & 28.8 \\
  PVTv2-B1~\cite{wang2022pvt} & 2.1 & 62.2 & 14.6 & 41.8 & 28.9 \\
  FastViT-T12~\cite{vasu2023fastvit} & 1.4 & 64.3 & 14.0 & 39.9 & 27.6 \\
  \rowcolor[gray]{0.92}
  \textbf{LSNet-B} & 1.3 & \textbf{59.3} & \textbf{17.3} & \textbf{43.1} & \textbf{30.7} \\
  \bottomrule
  \end{tabular}
  }
  \vspace{-20pt}
\end{table}

\subsection{Model Analyses}
\label{analyses:settings}
We conduct experiments to analyze the design elements in LSNet on ImageNet-1K. Following~\cite{graham2021levit,liu2023efficientvit}, all models are trained for 100 epochs for limitations in training time and computation resource. LSNet-T is employed for analyses, with $K_L=7$, $K_S=3$ and $C/G=8$, by default.

\textbf{Effectiveness of LS convolution.} We analyze the effectiveness of our proposed LS convolution by first comparing it with ``w/o LS conv.'', in which all LS convolutions are replaced with identity functions. As shown in \cref{tab:lsconv}, our LS convolution improves 2.3\% top-1 accuracy with only 0.02G FLOPs increase compared with ``w/o LS conv.''. Furthermore, we compare our LS convolution with other effective token mixing methods by directly replacing all LS convolutions with others. As shown in \cref{tab:lsconv}, LS convolution achieves superior performance with low computational costs. By employing other methods, the top-1 accuracy consistently decreases. Compared with (S)W-SA~\cite{liu2021swin}, SDTA~\cite{maaz2022edgenext}, and LSRA~\cite{wang2022pvt}, LS convolution obtains improvements of 0.8\%, 1.0\%, and 1.1\% top-1 accuracies, respectively, with fewer FLOPs. Besides, LS convolution outperforms RepMixer~\cite{vasu2023fastvit} and CGA~\cite{liu2023efficientvit} by 1.9\% and 1.1\% top-1 accuracies, respectively. Meanwhile, we compare our LS convolution with other dynamic convolutions by simply replacing the LS convolution. As shown in \cref{tab:dynamic}, thanks to incorporating large-field perception and small-field aggregation, LS convolution exhibits superiority in terms of accuracy and efficiency compared with other methods. For example, LS convolution surpasses CondConv~\cite{yang2019condconv} and DY-Conv~\cite{chen2020dynamic} by considerable margins of 1.8\% and 1.6\% top-1 accuracies, respectively, well showing the effectiveness.

\textbf{Importance of large-kernel perception.} We verify the effect of large-kernel perception (LKP) by first comparing it with ``w/o LKP'', in which we remove the large-kernel depth-wise convolution in the LKP. As shown in \cref{tab:lkp-ska}, we can observe that the top-1 accuracy is significantly reduced by 1.1\% in the absence of the large-field perception. We further investigate the impact of the large-kernel size, \ie, $K_L$, in the LKP. As shown in \cref{tab:lkp-ska}, the model performance continues to increase as the kernel size grows larger, showing the benefit of capturing contextual information with a large receptive field. Besides, the top-1 accuracy reaches a saturation point around a kernel size of 7, which is similar to the observations in previous works~\cite{liu2022convnet}.

\begin{figure}[!t]
\begin{minipage}{.23\textwidth}
  \captionof{table}{Superiority of LS conv.}
  \label{tab:lsconv}
  \centering
  \resizebox{1\columnwidth}{!}{
  \begin{tabular}{lcc}
    \toprule
    & FLOPs & Top-1 \\
    \toprule 
    w/o LS conv. & 0.29 & 69.3 \\
    \rowcolor[gray]{0.92}
    \textbf{LS conv.} & 0.31 & \textbf{71.6} \\
    \hline
    (S)W-SA~\cite{liu2021swin} & 0.36 & 70.8 \\ 
    SDTA~\cite{maaz2022edgenext} & 0.37 & 70.6 \\
    LSRA~\cite{wang2022pvt} & 0.37 & 70.5 \\
    RepMixer~\cite{vasu2023fastvit} & 0.29 & 69.7  \\ 
    CGA~\cite{liu2023efficientvit} & 0.32  & 70.5  \\
    AFF~\cite{huang2023adaptive} & 0.30 & 69.5 \\
    \bottomrule
  \end{tabular}
  }
\end{minipage}
\hfill
\begin{minipage}{.23\textwidth}
  \captionof{table}{Comparing other conv.}
  \label{tab:dynamic}
  \centering
  \resizebox{1\columnwidth}{!}{
  \begin{tabular}{lcc}
    \toprule
      & FLOPs & Top-1 \\
    \toprule 
    \rowcolor[gray]{0.92}
    \textbf{LS conv.} & 0.31 & \textbf{71.6} \\
    \midrule
    CondConv~\cite{yang2019condconv} & 0.29 & 69.8 \\
    DY-Conv~\cite{chen2020dynamic} & 0.29 & 70.0 \\
    Involution~\cite{li2021involution} & 0.31 & 70.3 \\
    DCD~\cite{li2021revisiting}  & 0.29 & 69.8 \\
    CoT~\cite{li2022contextual} & 0.37 & 71.1 \\
    ODConv~\cite{li2022omni} & 0.29 & 70.0 \\
    \bottomrule
  \end{tabular}
  }
\end{minipage}
\vspace{-5pt}
\end{figure}

\begin{figure}[!t]
\begin{minipage}{.23\textwidth}
  \captionof{table}{LKP and SKA.}
  \label{tab:lkp-ska}
  \centering
  \resizebox{1\columnwidth}{!}{
  \begin{tabular}{lcc}
    \toprule
      & FLOPs & Top-1 \\
    \toprule 
    \rowcolor[gray]{0.92}
    \textbf{LSNet-T} & 0.31 & \textbf{71.6} \\
    \hline
    w/o LKP & 0.31 & 70.5 \\ 
    $K_L= 3$ & 0.31 & 70.9 \\ 
    $K_L= 5$ & 0.31 & 71.2 \\
    $K_L= 9$ & 0.32 & 71.5 \\
    \hline
    w/o SKA & 0.31 & 70.1 \\ 
    $K_S= 1$  & 0.30 & 69.6 \\ 
    $K_S= 5$  & 0.34 & 71.6 \\
    \bottomrule
  \end{tabular}
  }
\end{minipage}
\hfill
\begin{minipage}{.23\textwidth}
  \captionof{table}{Other designs.}
  \label{tab:channel}
  \centering
  \resizebox{1\columnwidth}{!}{
  \begin{tabular}{lcc}
    \toprule
      & FLOPs & Top-1 \\
    \toprule 
    \rowcolor[gray]{0.92}
    \textbf{LSNet-T} & 0.31 & \textbf{71.6} \\
    \midrule
    $C/G=1$ & 0.38 & 71.7 \\ 
    $C/G=4$ & 0.33 & 71.6 \\ 
    $C/G=16$  & 0.31 & 71.3 \\ 
    $C/G=32$  & 0.31 & 70.9 \\ 
    \midrule
    w/o DW & 0.31 & 71.1 \\
    w/o SE & 0.31 & 71.3 \\
    \bottomrule
  \end{tabular}
  }
\end{minipage}
\vspace{-15pt}
\end{figure}

\textbf{Importance of small-kernel aggregation.} We show the importance of small-kernel aggregation (SKA) by first comparing it with ``w/o SKA'', in which we leverage a static depth-wise convolution with the kernel size of $K_S\times K_S$ to directly process the outcome of LKP as the output. Note that ``w/o SKA'' is the combination of large-kernel and small-kernel convolutions. \cref{tab:lkp-ska} presents the comparison results. We can observe that our LS convolution significantly outperforms ``w/o SKA'' by 1.5\% top-1 accuracy. It highlights the superiority of our LS convolution over the simple combination of large-kernel and small-kernel convolutions. Additionally, we inspect the impact of the contextual scope of aggregation, \ie, $\mathcal{N}_{K_S}(x_i)$, by adopting different $K_S$ in the SKA. As shown in \cref{tab:lkp-ska}, we can achieve the optimal trade-off between accuracy and computational costs under the $K_S$ of 3. It demonstrates the efficacy of adaptive aggregation in highly related surroundings.

\textbf{Impact of the number of groups.} We inspect the impact of different numbers of groups, \ie, $G$, in LS conv. As $G$ increases, the number of channels with shared aggregation weights, \ie, $\frac{C}{G}$, decreases, with higher computational costs. As shown in \cref{tab:channel}, as $\frac{C}{G}$ increases from 1 to 32, the top-1 accuracy decreases from 71.7\% to 70.9\%, along with the reduced computation complexity. It shows the benefit of performing different aggregation ways for varying channels, due to that they usually encode different representation subspaces and diverse semantic attributes~\cite{bau2020understanding}. Besides, we can observe that $\frac{C}{G}=8$ achieves the best balance.

\textbf{Impact of extra DW and SE layers.} We verify the effect of the extra depth-wise convolution and SE layer by removing them separately, which are denoted as ``w/o DW'' and ``w/o SE'', respectively. In \cref{tab:channel}, they decrease the top-1 accuracy by 0.5\% and 0.3\%, respectively, showing the efficacy of introducing more local structural information.

\textbf{Generalization of LS convolution to other architectures.} We show the generalization of LS convolution by transferring it to other vision networks. Specifically, we conduct experiments on two widely recognized architectures, \ie, ResNet~\cite{he2016deep} and DeiT~\cite{touvron2021training}, by simply replacing their all 3$\times$3 convolution, and self-attention with LS convolution, respectively. All models are 
\begin{table}
  \caption{Generalization ability of LS convolution on other architectures. We simply replace 3$\times$3 convolution and self-attention with LS convolution for ResNet and DeiT, respectively.}
  \label{tab:generalization}
  \centering
  \small
  \resizebox{0.8\columnwidth}{!}{
  \begin{tabular}{cccc}
  \toprule
  Model          & LS conv. &  FLOPs (G) & Top-1 (\%) \\
  \toprule
  ResNet50 & $\times$ & 4.1 & 78.8 \\
  ResNet50 & \checkmark & 2.6 & 80.7 \\
  DeiT-T & $\times$ & 1.3 & 72.2 \\
  DeiT-T & \checkmark & 0.9 & 73.0 \\
  \bottomrule
  \end{tabular}
  }
  \vspace{-15pt}
\end{table}
trained under the same settings for 300 epochs. As shown in \cref{tab:generalization}, incorporating LS convolution into ResNet50, and DeiT-T significantly improves their top-1 accuracies by 1.9\%, and 0.8\%, respectively, which showcases its good generalization capability.

\section{Conclusion}
In this work, we present LSNet, a novel family of lightweight vision networks that integrates the ``See Large, Focus Small'' strategy inspired by the human vision system. LSNet incorporates LS convolution, a new operation that combines large-kernel perception and small-kernel aggregation, enabling efficient and accurate processing of visual information. Extensive experiments demonstrate that LSNet achieves state-of-the-art performance and efficiency trade-offs. It shows the superiority over others across diverse tasks. We hope that LSNet can serve as a strong baseline and inspire further advancements in the development of lightweight and efficient vision networks.

\section{Acknowledgments}
This work was supported by Beijing Natural Science Foundation (Nos. L223023, L247026), National Natural Science Foundation of China (Nos. 62271281, 62441235, 62021002), and the Key R \& D Program of Xinjiang, China (2022B01006).

{
    \small
    \bibliographystyle{ieeenat_fullname}
    \bibliography{main}
}

\newpage

\appendix

\section{Implementation and Architectural Details}
\subsection{Implementation Details}
For image classification on ImageNet-1K~\cite{deng2009imagenet}, we adopt the same training recipe as~\cite{liu2023efficientvit,huang2023adaptive,vasu2023fastvit}. Specifically, we employ the standard image size of 224$\times$224 for both training and testing. All models are trained from scratch for 300 epochs. We use the AdamW optimizer~\cite{loshchilov2017decoupled} with a cosine learning rate scheduler. The initial learning rate is set to 4$\times$10$^{-3}$, and the total batch size is set to 2048. For data augmentation, we leverage mixup~\cite{zhang2017mixup}, RandAugment~\cite{cubuk2020randaugment}, CutMix~\cite{yun2019cutmix}, and random erasing~\cite{zhong2020random}, \etc. \cref{tab:hyperparam} provides the training details of LSNet.

For object detection and instance segmentation on COCO-2017~\cite{lin2014microsoft}, we employ the same training setting as~\cite{vasu2023fastvit,liu2023efficientvit,li2022efficientformer}. Specifically, we utilize the AdamW optimizer and train the model for 12 epochs with a batch size of 16. The training resolution is 1333$\times$800 and the initial learning rate is set to 2$\times$10$^{-4}$. The learning rate decays with a rate of 0.1 at the 8-th and 11-th epochs. We initialize the backbones with the pretrained ImageNet-1K weights.

For semantic segmentation on ADE20K~\cite{zhou2017scene}, following~\cite{li2023rethinking,vasu2023fastvit}, all models are trained for 40K iterations by the AdamW~\cite{loshchilov2017decoupled} optimizer with a batch size of 32. We adopt the poly learning rate schedule with the power of 0.9 and the initial learning rate of 2$\times$10$^{-4}$, like~\cite{li2023rethinking,vasu2023fastvit}. We employ the training resolution of 512$\times$512 and report the single scale testing results on the ADE20K validation set, as in~\cite{yu2022metaformer,pan2022edgevits}. The backbone models are initialized with the pretrained weights on ImageNet-1K.

For robustness evaluation, following~\cite{liu2022convnet,vasu2023fastvit,mao2022towards}, we employ the ImageNet-C~\cite{hendrycks2019benchmarking}, ImageNet-A~\cite{hendrycks2021natural}, ImageNet-R~\cite{hendrycks2021many}, and ImageNet-Sketch~\cite{wang2019learning} benchmarks. Specifically, ImageNet-C consists of algorithmically generated corruptions that are applied to the ImageNet test set. ImageNet-A contains naturally occurring examples misclassified by ResNets~\cite{he2016deep}. ImageNet-R comprises natural renditions of object classes in ImageNet, incorporating various textures and image statistics. ImageNet-Sketch includes white and black sketches of all ImageNet classes, gathered through google image queries.

\begin{table}[h]
    \caption{Training details on ImageNet-1K.}
    \label{tab:hyperparam}
    \small
    \centering
    \begin{tabular}{cc}
        \toprule
        Model & LSNet-T/S/B                \\
        \hline
        optimizer       & AdamW                 \\
        batch size      & 2048         \\
        training epochs & 300                   \\
        LR schedule     & cosine                \\
        learning rate   & 0.004        \\
        warmup epochs   & 5                     \\
        weight decay    & 0.025/0.025/0.05                 \\
        augmentation    & RandAug(9, 0.5)       \\
        random erase    & 0.25                  \\
        color jitter    & 0.4                   \\
        mixup           & 0.8                   \\
        cutmix          & 1.0                   \\
        gradient clip   & 0.02                  \\
        label smooth     & 0.1                   \\
        \bottomrule
    \end{tabular}
\end{table}

\begin{table}[h]
    \caption{Architectural details of LSNet variants.}
    \label{tab:achitecture}
    \small
    \centering
    \resizebox{\columnwidth}{!}{%
    \begin{tabular}{c|c|c|c|c|c|c}
    \toprule
    \multirow{2}{*}{Stage} & \multirow{2}{*}{Resolution} &    \multirow{2}{*}{Type} & \multirow{2}{*}{Config} & \multicolumn{3}{c}{LSNet}  \\ \cline{5-7}
    & & & & T & S & B \\
    \hline
    \hline
    \multirow{6}{*}{stem} & \multirow{2}{*}{$\frac{H}{2} \times \frac{W}{2}$} & \multirow{2}{*}{Convolution} & \multirow{2}{*}{channels} & \multirow{2}{*}{16} & \multirow{2}{*}{24} & \multirow{2}{*}{32} \\
    & & & & & \\
    \cline{2-7}
    & \multirow{2}{*}{$\frac{H}{4} \times \frac{W}{4}$} & \multirow{2}{*}{Convolution} & \multirow{2}{*}{channels} & \multirow{2}{*}{32} & \multirow{2}{*}{48} & \multirow{2}{*}{64} \\
    & & & & & \\
    \cline{2-7}
    & \multirow{2}{*}{$\frac{H}{8} \times \frac{W}{8}$} & \multirow{2}{*}{Convolution} & \multirow{2}{*}{channels} & \multirow{2}{*}{64} & \multirow{2}{*}{96} & \multirow{2}{*}{128} \\
    & & & & & \\
    \hline
    \multirow{2}{*}{1} & \multirow{2}{*}{$\frac{H}{8} \times \frac{W}{8}$} & \multirow{2}{*}{LS Block} & channels & 64 & 96 & 128 \\
    \cline{4-7}
    & & & blocks & 0 & 1 & 4 \\
    \hline
    \multirow{2}{*}{2} & \multirow{2}{*}{$\frac{H}{16} \times \frac{W}{16}$} & \multirow{2}{*}{LS Block} & channels & 128 & 192 & 256 \\
    \cline{4-7}
    & & & blocks & 2 & 2 & 6 \\
    \hline
    \multirow{2}{*}{3} & \multirow{2}{*}{$\frac{H}{32} \times \frac{W}{32}$} & \multirow{2}{*}{LS Block} & channels & 256 & 320 & 384 \\
    \cline{4-7}
    & &  & blocks & 8 & 8 & 8 \\
    \hline
    \multirow{2}{*}{4} & \multirow{2}{*}{$\frac{H}{64} \times \frac{W}{64}$} & \multirow{2}{*}{MSA Block} & channels & 384 & 448 & 512 \\
    \cline{4-7}
    & & & blocks & 10 & 10 & 10 \\
    \bottomrule
    \end{tabular}
    }
    \vspace{-10pt}
\end{table}

\subsection{Architectural Details}
\cref{tab:achitecture} presents the architectural details of LSNet variants, which are distinguished by the number of blocks and the number of channels within each stage. 

\section{More Comparisons}
We present more comparisons between LS convolution and others from mathematical perspectives. Specifically, for simply combining large-kernel with small-kernel convolutions, it follows the similar perception $\mathcal{P}_{conv}$ and aggregation $\mathcal{A}_{conv}$ processes as the standard convolution, \ie, leveraging relative positions for relationship modeling and static kernel weights for feature integration. However, compared with LS convolution, it suffers from the limited modeling capability due to the lack of adaptability for different contexts. In other dynamic ways, Involution~\cite{li2021involution} leverages MLP for perception $\mathcal{P}_{inv}$ to derive the aggregation weights conditioned on $x_i$. Its aggregation $\mathcal{A}_{inv}$ then use the weights to convolve the features in $\mathcal{N}_{K}(x_i)$ with the process of $y_i = \mathcal{A}_{inv}(\mathcal{P}_{inv}(x_i), \mathcal{N}_{K}(x_i)) = \text{MLP}(x_i) \circledast \mathcal{N}_{K}(x_i)$. Although the aggregation process is dynamic, its perception process is confined to $x_i$, which leads to inadequate neighborhood relationship modeling compared with LS convolution. Additionally, CondConv~\cite{yang2019condconv} proposes per-example routing with global average pooling and MLP to linearly combining multiple convolution kernels for the aggregation weights in its perception $\mathcal{P}_{cond}$. Its aggregation $\mathcal{A}_{cond}$ then convolves the features in $\mathcal{N}_{K}(x_i)$ with the weights. Its process $y_i = \mathcal{A}_{cond}(\mathcal{P}_{cond}(X), \mathcal{N}_{K}(x_i))$ can be formulated as $y_i = (\textstyle \sum \text{MLP}(\text{GAP}(X))\cdot W_{cond}) \circledast \mathcal{N}_{K}(x_i)$. However, unlike LS convolution, CondConv leverages example-dependent perception, which prevents distinct tokens to adapt to diverse contexts.

\section{Qualitative Analyses}
\subsection{Analyses for LS Convolution}

\begin{figure}[t]
\centering
    \includegraphics[width=0.9\linewidth]{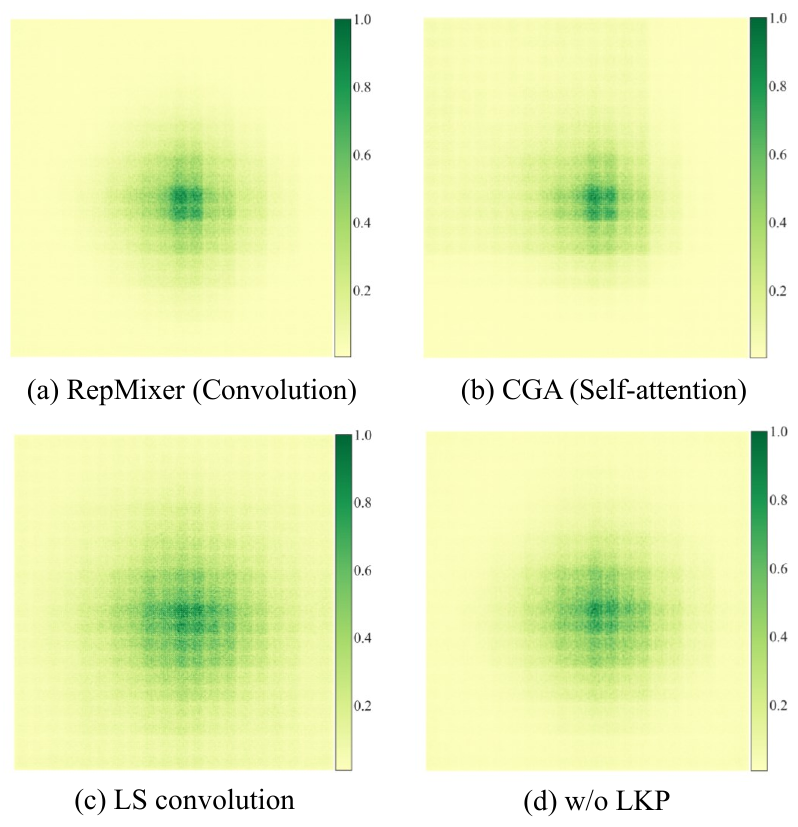}
    \caption{Visualization of the effective receptive field. Best viewed when zoomed in. (a) and (b) show that RepMixer and CGA exhibit unnatural patterns in the effective receptive field. (c) illustrates that LS convolution enables broad peripheral perception and central view focusing simultaneously. (d) shows that without LKP, LS convolution presents a smaller receptive field compared with (c), indicating the effectiveness of LKP.}
    \label{fig:erf}
    \vspace{-10pt}
\end{figure}

\begin{figure*}[t]
\centering
    \includegraphics[width=0.9\linewidth]{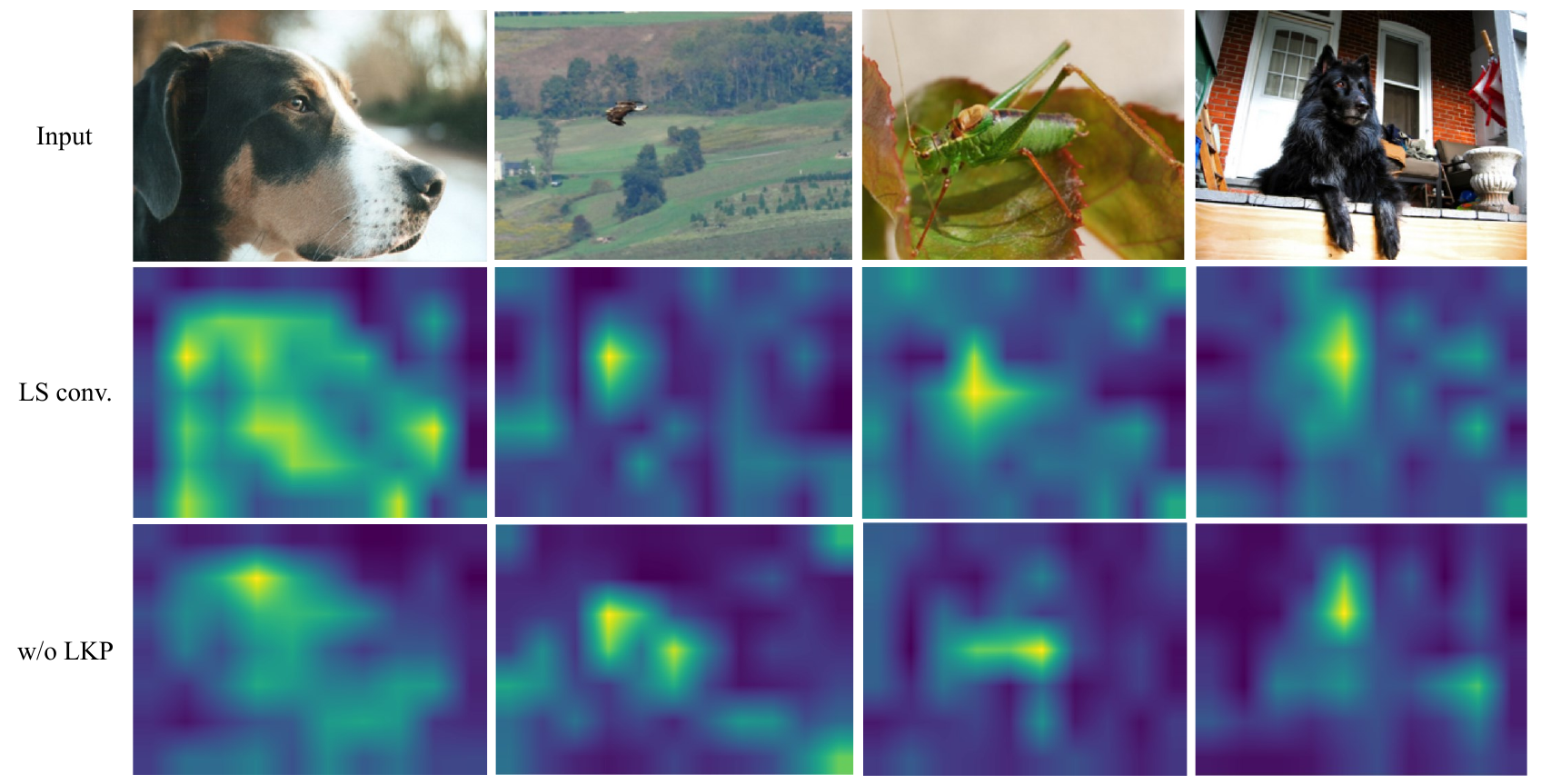}
    \caption{Visualization of the aggregation weights in LS convolution. The second row shows that the aggregation weights are well correlated with semantic relevant areas. The third row indicates that integrating LKP enables LS convolution to capture more precise visual patterns with improved contextual information.}
    \label{fig:ls}
\end{figure*}

\begin{figure*}[t]
\centering
    \includegraphics[width=0.9\linewidth]{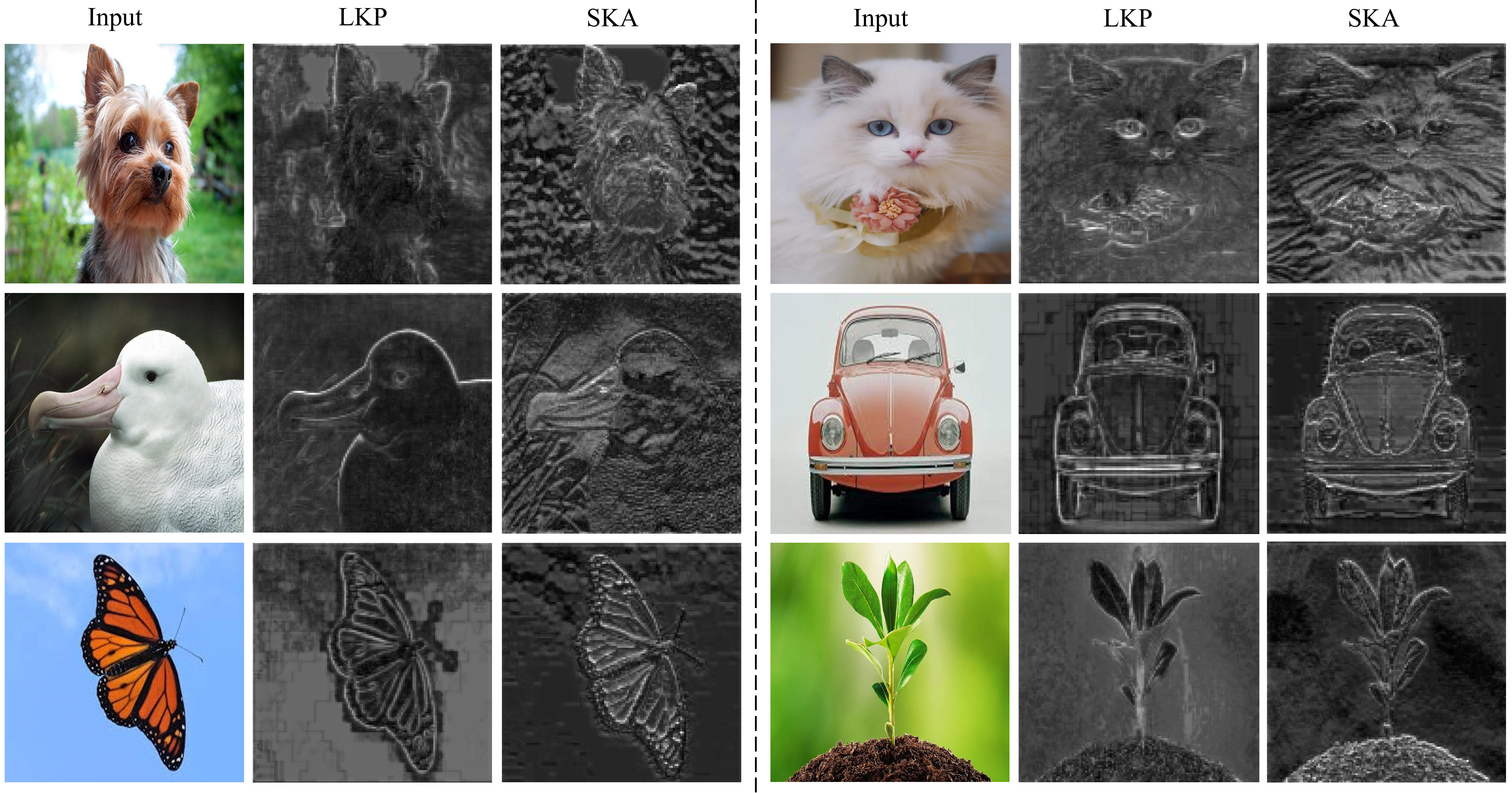}
    \caption{Visualization of the feature maps of LKP and SKA. The second column in each part shows that LKP can encompass a broad view of the scene. The third column in each part indicates that based on LKP, SKA can further grasp more subtle features and detailed patterns.}
    \label{fig:feature}
\end{figure*}

\begin{figure*}[t]
\centering
    \includegraphics[width=0.9\linewidth]{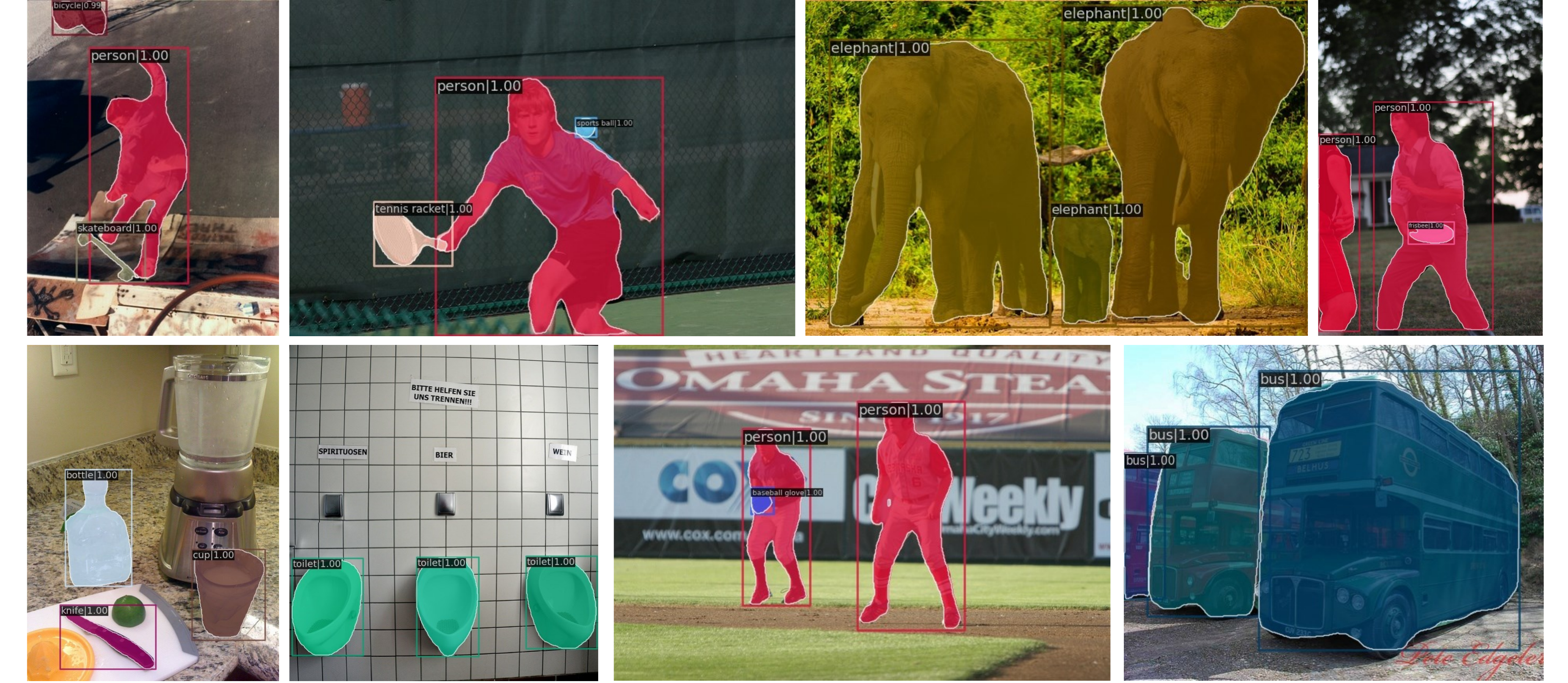}
    \caption{Qualitative results for object detection and instance segmentation on COCO-2017~\cite{lin2014microsoft}.}
    \label{fig:coco}
\end{figure*}

\begin{figure*}[t]
\centering
    \includegraphics[width=0.9\linewidth]{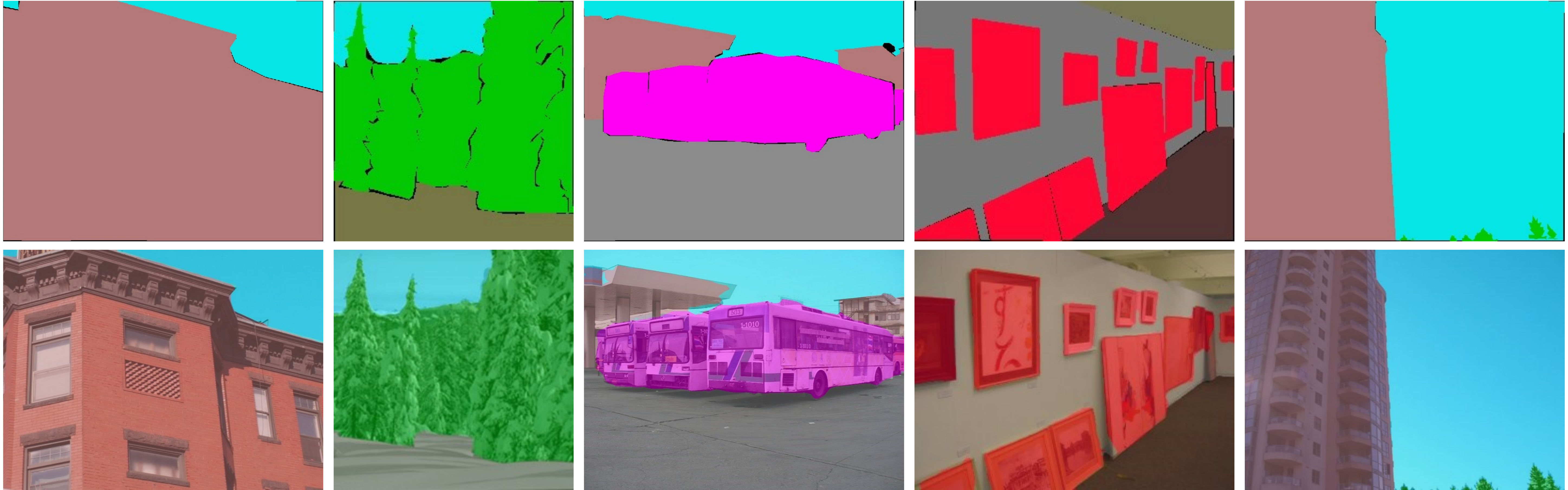}
    \caption{Qualitative results for semantic segmentation on ADE20K~\cite{zhou2017scene}. The upper row shows the ground truth masks, and the lower row presents the predicted masks.}
    \label{fig:ade}
\end{figure*}

We present the visualization analyses to qualitatively show the effectiveness of LS convolution. Specifically, we employ the effective receptive field~\cite{ding2022scaling,luo2016understanding} method to compare LS convolution with convolution and self-attention, based on LSNet-T. We introduce the state-of-the-art RepMixer~\cite{vasu2023fastvit} and CGA~\cite{liu2023efficientvit} as the representatives of convolution and self-attention, respectively. Besides, we simply replace all LS convolutions in the model with others. As shown in \cref{fig:erf}, RepMixer and CGA suffer from the unnatural patterns, caused by static convolution kernels and window-based self-attention, respectively. In contrast, LS convolution enjoys both central area focusing and extensive peripheral viewing, showing smooth visual processing. Meanwhile, compared with ``w/o LKP'' where the large-kernel depth-wise convolution in the LKP is removed, LS convolution exhibits an enlarged effective receptive field. It is attributed to the ability of LKP to efficiently capture broad contextual information. 

Furthermore, we conduct visualization for the aggregation weights in LS convolution. Specifically, we obtain the cumulative value of the aggregation coefficients corresponding to each token in all aggregation processes it is involved in. We then visualize the average of the absolute values of all channels in the last layer at the third stage and perform upsampling for display. As shown in \cref{fig:ls}, the aggregation weights of SKA enjoy favorable interpretability. They effectively strengthen semantically relevant vision regions and accurately capture discriminative patterns in images. Besides, compared with ``w/o LKP'', LS convolution exhibits more precise emphasis on important visual areas, showcasing the improved modeling of spatial relationships facilitated by LKP. Based on LKP and SKA, LS convolution can thus help the model to grasp the critical visual information under limited computational costs, enhancing both efficiency and effectiveness.

Besides, we also visualize the feature maps generated by the LKP and SKA for more inspection. Specifically, we use the features after the large-kernel depth-wise convolution and the small-kernel dynamic convolution in the first stage for demonstration. As shown in \cref{fig:feature}, the feature maps produced by LKP exhibit a broad receptive field, capturing a wide range of contextual information in the scene. This characteristic is reminiscent of the human peripheral vision system, adept at sensing the general surroundings. On the other hand, based on LKP, SKA further demonstrates the ability to grasp finer details within the image. It can result in more subtle features like gradients of hairs and clear outlines. This behavior is analogous to the human central vision system, which excels at discerning fine details and high-resolution information. Thanks to them, LS convolution can well help the model achieve the effective and efficient perception and aggregation processes.

\subsection{Analyses for Downstream Tasks}
We present the qualitative results when integrating LSNet into the Mask-RCNN framework~\cite{he2017mask} for object detection and instance segmentation tasks, and into the Semantic FPN framework~\cite{kirillov2019panoptic} for the semantic segmentation task. As illustrated in \cref{fig:coco}, the model can achieve precise detection and segmentation of instances in diverse images. Besides, as shown in \cref{fig:ade}, the model demonstrates the ability to generate high-quality semantic segmentation masks.

\section{Contribution, Limitation, and Impact}
\textbf{Contribution.} In summary, our contributions are threefold, as follows:
\begin{enumerate}
\item We advocate a new strategy ``See Large, Focus Small'', inspired by the human vision system, for lightweight and efficient network design. By encompassing a broad perceptual range with enriched contextual information, it facilitates focused feature aggregation, fostering detailed visual understanding.
\item We propose LS convolution as a novel operation for modeling visual features in lightweight models. LS convolution integrates large-kernel perception and small-kernel aggregation, enabling proficient processing of visual information through both effective and efficient perception and aggregation processes.
\item We present a new family of lightweight vision networks, namely LSNet, which is built on LS convolution. Extensive experiments demonstrate that LSNet achieves the state-of-the-art performance and efficiency trade-offs compared with other lightweight networks across a broad range of vision tasks.
\end{enumerate}

\textbf{Limitation.} Due to the limited computational resources, we do not extend the application of our LSNet to other scenarios, such as visual-language tasks or unsupervised learning. We do not investigate the pretraining of LSNet on large-scale datasets, \eg, ImageNet-21K~\cite{deng2009imagenet}, due to the same reason. However, we are enthusiastic about exploring more applications for LSNet in the future. 

\textbf{Societal Impact.} We observe that this study is purely academic, and we have not identified any direct negative social impact resulting from our work. Nevertheless, we acknowledge the potential for malicious use of our models, which is a concern that affects the field. While we believe that it should be mitigated, discussions concerning this matter are beyond the scope of this paper.

\end{document}